\documentclass[a4paper,UKenglish,cleveref, autoref, thm-restate]{lipics-v2021}
\usepackage{xspace}
\usepackage{tikz}
\usepackage[ruled,linesnumbered,lined,boxed,commentsnumbered]{algorithm2e}
\usepackage{import}
\hideLIPIcs
\nolinenumbers 

\title{Solver-Aided Expansion of Loops to Avoid Generate-and-Test}
\titlerunning{Solver-Aided Expansion of Loops to Avoid Generate-and-Test}

\author{Niklas {Dewally}}{School of Computer Science, University of St Andrews, UK \and \url{https://niklas.dewally.com/}}{niklas@dewally.com}{https://orcid.org/0009-0004-6662-020X}{}

\author{Özgür {Akgün}}{School of Computer Science, University of St Andrews, UK \and \url{https://www.st-andrews.ac.uk/computer-science/people/oa86}}{ozgur.akgun@st-andrews.ac.uk}{https://orcid.org/0000-0001-9519-938X}{}

\authorrunning{N, Dewally, Ö. Akgün}


\acknowledgements{We thank the members of the ``AI for Decision Making'' Vertically Integrated Project at the University of St Andrews for their contributions.}

\newcommand{\MiniZinc}{\textsc{MiniZinc}\xspace}
\newcommand{\EssencePrime}{\textsc{Essence Prime}\xspace}
\newcommand{\SavileRow}{\textsc{Savile Row}\xspace}
\newcommand{\Conjure}{\textsc{Conjure}\xspace}
\newcommand{\Essence}{\textsc{Essence}\xspace}
\newcommand{\Minion}{\textsc{Minion}\xspace}

\supplement{Our prototype extension to \Conjure is available at \url{https://github.com/conjure-cp/conjure-oxide}. The models we use in our experiments, scripts we use to run the experiments, raw results and plotting scripts are available at \url{https://github.com/niklasdewally/2025-comprehension-unrolling-modref}.}
\supplementdetails[subcategory={Source Code}]{Software}{https://doi.org/10.5281/zenodo.16723727}
\supplementdetails{Dataset}{https://doi.org/10.5281/zenodo.16724771}

\EventEditors{John Q. Open and Joan R. Access}
\EventNoEds{2}
\EventLongTitle{42nd Conference on Very Important Topics (CVIT 2016)}
\EventShortTitle{CVIT 2016}
\EventAcronym{CVIT}
\EventYear{2016}
\EventDate{December 24--27, 2016}
\EventLocation{Little Whinging, United Kingdom}
\EventLogo{}
\SeriesVolume{42}
\ArticleNo{23}

\begin{document}

\maketitle

\begin{abstract}
Constraint modelling languages like \MiniZinc and \Essence rely on unrolling loops (in the form of quantified expressions and comprehensions) during compilation.
Standard approaches generate all combinations of induction variables and use partial evaluation to discard those that simplify to identity elements of associative-commutative operators (e.g. true for conjunction, 0 for summation).
This can be inefficient for problems where most combinations are ultimately irrelevant.
We present a method that avoids full enumeration by using a solver to compute only the combinations required to generate the final set of constraints.
The resulting model is identical to that produced by conventional flattening, but compilation can be significantly faster.
This improves the efficiency of translating high-level user models into solver-ready form, particularly when induction variables range over large domains with selective preconditions.
\end{abstract}


\section{Introduction\label{intro}}

Constraint modelling languages such as \Essence~\cite{akgun2022conjure}, \EssencePrime~\cite{nightingale2017automatically}, and \MiniZinc~\cite{nethercote2007minizinc} allow users to specify complex combinatorial problems in a concise and declarative manner. However, the efficiency of compiling these models often hinges on how loops and quantified expressions are expanded, minor syntactic changes can lead to substantial differences in compilation time and memory use. Current tools rely on generate-and-test strategies combined with partial evaluation, making performance fragile and dependent on modeller expertise.

We present a solver-aided method for expanding comprehensions and quantified expressions that is robust to syntactic formulation. Our approach eliminates the need for full enumeration and partial evaluation, ensuring that only relevant combinations of induction variables are generated, regardless of how guards or conditions are expressed. This step moves declarative modelling closer to its ideal: enabling users to express intent without hidden performance traps. The underlying technique is general and could be adapted for any context where side-effect-free loop unrolling is required.

Empirical results on standard benchmarks demonstrate that our solver-aided expansion method robustly scales across different syntactic formulations of constraints. It significantly reduces compilation times compared to traditional generate-and-test approaches, particularly when implicit conditions or large domains are involved. This robustness and efficiency makes constraint model compilation more reliable, allowing users to express intent clearly without concern for hidden performance costs.


\section{Motivation}

Efficient unrolling of loops, quantified expressions, and comprehensions is critical in constraint modelling languages such as \MiniZinc and \EssencePrime, as it directly affects the scalability and performance of model compilation. Typically, these languages expand loops into explicit constraints through generate-and-test strategies, enumerating all combinations and subsequently filtering irrelevant cases.

\begin{figure}
  \centering
  \captionsetup[subfigure]{justification=centering}
  \begin{subfigure}[t]{\textwidth}
    \centering
    \begin{lstlisting}[]
int: n;
array[1..n] of var 1..2: class;
constraint
  forall(a,b,c in 1..n)(
    (a<=b /\ b<=c /\ a^2 + b^2 = c^2) -> (
      class[a] != class[b] \/ 
      class[b] != class[c] \/
      class[c] != class[a]));
solve satisfy;
    \end{lstlisting}
    \caption{Naive \MiniZinc model. As an indicative figure, this model takes $7.8$ seconds to translate, with $n=200$.}
    \label{lst:minizinc-forall}
  \end{subfigure}

  \begin{subfigure}[t]{\textwidth}
    \centering
    \begin{lstlisting}[]
int: n;
array[1..n] of var 1..2: class;
constraint
  forall(a,b,c in 1..n)(
    if (a<=b /\ b<=c /\ a^2 + b^2 = c^2)
    then (class[a] != class[b] \/ 
          class[b] != class[c] \/
          class[c] != class[a])
    else true endif);
solve satisfy;
    \end{lstlisting}
    \caption{Another \MiniZinc model using an if statement. As an indicative figure, this model takes $15.3$ seconds to translate, with $n=200$.}
    \label{lst:minizinc-if}
  \end{subfigure}

  \begin{subfigure}[t]{\textwidth}
    \centering
    \begin{lstlisting}[]
int: n;
array[1..n] of var 1..2: class;
constraint
  forall(x in [ class[a] != class[b] \/
                class[b] != class[c] \/
                class[c] != class[a]
              | a,b,c in 1..n
                where (a<=b /\ b<=c /\ a^2 + b^2 = c^2)
              ])(x);
solve satisfy;
    \end{lstlisting}
    \caption{Faster \MiniZinc model using a static comprehension guard. As an indicative figure, this model takes $1.3$ seconds to translate, with $n=200$.}
    \label{lst:minizinc-comp}
  \end{subfigure}
  \caption{\MiniZinc models of the Boolean Pythagorean Triples Problem}
  \label{fig:minizinc-models}
\end{figure}

To illustrate, consider the Boolean Pythagorean Triples Problem \cite{BooleanPythagTriples}, a well-known constraint satisfaction problem that partitions integers into two classes such that no Pythagorean triple \texttt{(a,b,c)} lies entirely within a single class. A natural \MiniZinc formulation of this problem (\autoref{lst:minizinc-forall}) explicitly loops through all possible triples \texttt{(a,b,c)}, creating constraints for each combination. Although intuitive, this approach scales poorly: \MiniZinc takes approximately 1.07 seconds for $n=100$ and dramatically increases to 7.82 seconds at $n=200$, due to enumerating and evaluating all $n^3$ combinations. Using an if statement instead of a logical implication as in \autoref{lst:minizinc-if} roughly doubles the time taken.

In contrast, adding a simple comprehension guard drastically improves \MiniZinc{}'s compilation time. \autoref{lst:minizinc-comp}, which filters combinations explicitly through a comprehension guard, compiles significantly faster (1.28 seconds for $n=200$), despite being semantically equivalent to the original formulation. This performance discrepancy highlights how subtle modelling differences critically impact practical usability.

\begin{figure}
  \centering
  \captionsetup[subfigure]{justification=centering}
  \begin{subfigure}[t]{\textwidth}
    \centering
    \begin{lstlisting}[language=Python]
triples = []
for a in range(1, n+1):
  for b in range(1, n+1):
    for c in range(1, n+1):
      if a <= b <= c and a ** 2 + b ** 2 == c ** 2:
        triples.append([a,b,c])
    \end{lstlisting}
    \caption{Naive Python implementation. For $n=1000$, this program takes roughly $3$ minutes to run.}
    \label{lst:pythontriples}
  \end{subfigure}

  \begin{subfigure}[t]{\textwidth}
    \centering
    \begin{lstlisting}[language=Python]
triples = []
for a in range(1, n+1):
  for b in range(a, n+1):
    c_squared = a**2 + b**2
    c = int(c_squared ** 0.5)
    if c < b or c > n:
      continue
    if c**2 == c_squared:
        triples.append([a, b, c])
    \end{lstlisting}
    \caption{Optimised Python implementation. For $n=1000$, this program takes less than a second to run.}
    \label{lst:pythontriplesfaster}
  \end{subfigure}
  \caption{Python enumeration of the relevant triples for the Boolean Pythagorean Triples Problem}
  \label{fig:python-models}
\end{figure}

Such issues are not exclusive to constraint modelling languages. A naive implementation of triple generation in Python (\autoref{lst:pythontriples}) also enumerates all possibilities and performs poorly, taking roughly two minutes for $n=1000$. While an optimised Python version (\autoref{lst:pythontriplesfaster}) is faster, it requires manual optimisation and increased complexity, contradicting the simplicity and clarity benefits provided by declarative languages.

\begin{figure}
  \centering
  \captionsetup[subfigure]{justification=centering}
  \begin{subfigure}[t]{\textwidth}
    \centering
    \begin{lstlisting}
given n: int(1..)
find class: matrix indexed by [int(1..n)] of int(1..2)
such that
  forAll a,b,c: int(1..n). 
    (a**2 + b**2 = c**2 /\ a<=b /\ b<=c) -> 
    or([class[a] != class[b],
        class[b] != class[c],
        class[c] != class[a]]) 
    \end{lstlisting}
    \caption{Using a \texttt{forall} quantified expression. As an indicative figure, this model takes $9.4$ seconds to translate, with $n=200$.}
    \label{lst:triples-eprime-forall}
  \end{subfigure}

  \begin{subfigure}[t]{\textwidth}
    \centering
    \begin{lstlisting}
given n: int(1..)
find class: matrix indexed by [int(1..n)] of int(1..2)
such that
  and([or([ class[a] != class[b]
          , class[b] != class[c]
          , class[c] != class[a]])
      | a,b,c: int(1..n), a<=b, b<=c, a**2+b**2=c**2
      ])
    \end{lstlisting}
    \caption{Using a comprehension with static guards. As an indicative figure, this model takes $11.5$ seconds to translate, with $n=200$.}
    \label{lst:triples-eprime-compcond}
  \end{subfigure}

  \begin{subfigure}[t]{\textwidth}
    \centering
    \begin{lstlisting}
given n: int(1..)
find class: matrix indexed by [int(1..n)] of int(1..2)
such that
  and([(a**2 + b**2 = c**2 /\ a<=b /\ b<=c) ->
       or([class[a] != class[b]
          , class[b] != class[c]
          , class[c] != class[a]]) 
      | a,b,c: int(1..n)
      ])
    \end{lstlisting}
    \caption{Using a comprehension without static guards. The translation of this model timed out in an hour, with $n=200$.}
    \label{lst:triples-eprime-inreturnexpr}
  \end{subfigure}

  \caption{\EssencePrime models of the Boolean Pythagorean Triples Problem}
  \label{fig:eprime-models}
\end{figure}

\EssencePrime, the input language of the modelling tool \SavileRow, faces similar performance concerns. A \EssencePrime formulation using comprehensions without guards (\autoref{lst:triples-eprime-inreturnexpr}) compiles inefficiently (162.43 seconds for $n=100$), as all combinations are generated and then filtered using partial evaluation. Conversely, a guarded comprehension (\autoref{lst:triples-eprime-compcond}) compiles notably faster (1.95 seconds for $n=100$), as does a direct formulation using quantifiers (\autoref{lst:triples-eprime-forall}, 5.85 seconds for $n=100$). These differences are critical, especially as automated model-generation tools like Conjure frequently produce formulations without explicit guards.

Current methods relying on partial evaluation suffer from substantial overhead, particularly in memory consumption and unnecessary computation, severely restricting scalability. Moreover, these methods are highly sensitive to subtle changes in model formulation, placing undue responsibility on modellers to optimise their syntax manually.

Our goal is to develop a robust method that eliminates reliance on generate-and-test and partial evaluation strategies. Instead, we propose using solver-aided expansion to compute only necessary combinations, ensuring consistent, efficient compilation regardless of how loops and comprehensions are expressed.


\section{Comprehensions in \EssencePrime}

Many constraint modelling languages, including \MiniZinc{} and \EssencePrime{}, provide \emph{comprehensions}—constructs that allow the concise specification of collections of expressions. The semantics of comprehensions in these languages are closely related: both allow modellers to describe large numbers of similar constraints in a declarative form, often using guards to restrict the range of assignments to induction variables. For this reason, we describe comprehensions as implemented in \EssencePrime{}; all key points apply to \MiniZinc{} as well, and our methods can be transferred without loss of generality.

\subsection{Syntax and Semantics}

In \EssencePrime{}, the general form of a matrix comprehension is:
\begin{verbatim}
    [return_expression | i: domain1, j: domain2, cond1, cond2, ...]
\end{verbatim}

Here, \emph{induction variables} (e.g., \texttt{i}, \texttt{j}) are introduced with their respective domains, and any subsequent expressions separated by commas are \emph{comprehension guards}, Boolean conditions restricting which assignments of the induction variables are included. Only assignments for which all guards hold are considered. The \emph{return expression} specifies the value to be computed for each such assignment.

For example:
\begin{verbatim}
    [ b - a | a: int(1..3), b: int(1..6), b % 3 = 0 ]
\end{verbatim}

This comprehension generates a matrix of the values \texttt{b - a} for all $a \in \{1,2,3\}$, $b \in \{1,\dots,6\}$, where $b$ is a multiple of 3. The expansion is limited to those combinations by the guard \texttt{b \% 3 = 0}.

In both \EssencePrime{} and \MiniZinc{}, comprehensions can be used to succinctly express conjunctions, disjunctions, sums, products, and more, by placing the comprehension within the appropriate aggregate operator (e.g., \texttt{and}, \texttt{or}, \texttt{sum}).

\subsection{Static and Dynamic Expressions}

A key distinction in the context of comprehension unrolling is between \emph{static} and \emph{dynamic} expressions:
\begin{itemize}
\item \textbf{Static expressions} reference only induction variables. These can be evaluated entirely at model compilation time.
\item \textbf{Dynamic expressions} involve non-induction variables (e.g., model variables or decision variables), whose values are only determined during solving.
\end{itemize}
Comprehension guards must be static, as they determine which assignments are included when unrolling the comprehension at compile time. Conditions that depend on decision variables cannot be used as guards and must be placed within the return expression.

For instance, suppose \texttt{m} and \texttt{n} are model variables. To specify that \texttt{m[i] = n[i]} only if \texttt{m[i]} is even, we write:
\begin{verbatim}
    and([ (m[i] % 2 = 0) -> m[i] = n[i] | i: int(1..n) ])
\end{verbatim}
Here, \texttt{m[i] \% 2 = 0} is a dynamic condition and appears inside the return expression rather than as a comprehension guard.

\subsection{Quantified Expressions}

\EssencePrime also includes quantified expressions, such as \texttt{forAll} and \texttt{exists}, which provide an alternative to comprehensions for expressing universal or existential constraints. The quantifier \texttt{forAll a, b: int(1..n). P(a, b)} can be expressed equivalently as:
\begin{verbatim}
    and([ P(a, b) | a: int(1..n), b: int(1..n) ])
\end{verbatim}

Similarly, existential quantification corresponds to an \texttt{or} of a comprehension. Unlike comprehensions, quantifiers do not support explicit comprehension guards; guards must be incorporated into the quantified condition itself. From a practical standpoint, quantifiers and comprehensions are interchangeable, as quantifiers are internally lowered to comprehensions during model rewriting in many systems, including \Conjure{}.

\subsection{Impact of Formulation on Rewriting Performance}

Although comprehensions with static guards, comprehensions with guards embedded in the return expression, and quantified expressions are semantically equivalent, only the first can be efficiently unrolled. When guards are omitted or moved inside the return expression, the model compiler is forced to enumerate all possible combinations of induction variables, then use partial evaluation to discard those which turn out to be irrelevant. For large domains, this generates significant memory and computational overhead, and can quickly become infeasible.

For example, consider:
\begin{verbatim}
    and([ (i % 2 = 0) -> m[i] = i | i: int(1..4) ])
\end{verbatim}

This comprehension unrolls to four expressions, of which only two matter after partial evaluation. For larger domains, the intermediate blowup can be severe.

In contrast, a comprehension guard such as \texttt{i \% 2 = 0} restricts the enumeration to only those values where the condition holds, directly controlling the size of the expansion and the resources required.

In summary, both the syntax and placement of conditions in comprehensions have a profound impact on the performance of model compilation. Small syntactic differences can determine whether the compiler must enumerate an infeasible number of combinations, or efficiently target only those that matter.

The next section describes our method for eliminating this fragility, making efficient expansion possible regardless of how conditions are written in the original model.


\section{Method: Solver-Aided Expansion of Loops}
\label{sec:solver-aided}

 
The scalability of constraint model compilation critically depends on efficient expansion of comprehensions and quantified expressions. In current systems, implicit or poorly-positioned static conditions often result in inefficient generate-and-test enumeration. We present a solver-aided approach that systematically extracts static conditions to ensure only necessary combinations of induction variables are generated, irrespective of the original syntactic form.

\subsection{Comprehension Unrolling as a Constraint Problem}

We reformulate comprehension unrolling as a constraint satisfaction problem,
referred to as a \textit{generator model}. Rather than enumerating all possible
assignments to induction variables and filtering them afterwards, we solve a
constraints model that finds assignments to the induction variables satisfying
the static guards of the comprehension.

Consider the following comprehension:

\begin{verbatim}
    and([ m[i] = i | i: int(1..4), i % 2 = 0 ])
\end{verbatim}

The induction variable \texttt{i} ranges over $\{1,2,3,4\}$, with the static guard \texttt{i \% 2 = 0}. The generator model is:

\begin{verbatim}
    find i: int(1..4)
    such that i % 2 = 0
\end{verbatim}

Solving this gives the assignments \texttt{i=2} and \texttt{i=4}, producing
the constraints \texttt{m[2]=2} and \texttt{m[4]=4}.

\subsection{Lifting Static Guards from Return Expressions}

For simpler comprehensions where static guards are explicitly defined as
comprehension guards, it is sufficient to use the comprehension guards as the
generator model, as described above. However, this approach relies on the user
to formulate the model in one of many possible ways, which we want to avoid. For
example, consider:

\begin{verbatim}
    and[(i % 2 = 0) -> m[i] = i | i: int(1..4)]
\end{verbatim}

A naive approach to finding the static guards would be to look for implications
inside \texttt{and} quantifiers, add the antecedents, if static, to the
generator model, and replace the return expression with the consequent. However,
this is not sufficient when static and dynamic guards are combined, such as in:

\begin{verbatim}
    and([!(i % 2 = 0 /\ m[i] % 2 = 0) -> m[i] = i | i: int(1..4)])
\end{verbatim}

To solve the problem of identifying static guards in the general case, we
convert the entire return expression into a static guard by replacing dynamic
sub-expressions with \textit{dummy variables}.

For example, we convert the return type of the above comprehension to a static
guard by replacing the dynamic expressions \texttt{m[i] \% 2 = 0} and
\texttt{m[i] = i} with boolean variables \texttt{Z1} and \texttt{Z2}, creating
the following generator model:

\begin{verbatim}
    find i: int(1..4)
    find Z1, Z2: bool
    branching on [i]
    such that 
      !(i % 2 = 0 /\ Z1) -> Z2 != true 
\end{verbatim}

As \texttt{Z1} and \texttt{Z2} are unconstrained, the solver enumerates all
possible values of \texttt{i} for which the original return expression is false,
for some values of \texttt{Z1} and \texttt{Z2}. As we do not care about the
values of the dummy variables, we only branch on the induction values. In
\EssencePrime, the semantics of the branching on statement means that this
generator model returns only unique assignments of the induction variables.

\subsection{Aggregates and Identity Values}

Aggregate comprehensions (e.g. \texttt{and}, \texttt{or}, \texttt{sum},
\texttt{product}) have known identity elements. The identity element can always
be removed from an operation's operands without affecting its result; thus, when
expanding comprehensions, we exclude expressions that we statically know to be
equal to the identity. To do this, in the generator model we ensure that the
rewritten return expression is not equal to the identity value of the
comprehension. For the example above, the generator model produces the values of
\texttt{i} where the result of \texttt{!(i \% 2 = 0 /\textbackslash\ m[i] \% 2 =
0)} depends on \texttt{m}.

\subsection{Algorithm\label{subsec:thealg}}

\begin{algorithm}[htp]
\caption{Substitute dynamic sub-expressions for dummy variables}
\label{alg:rewrite}
\SetKwData{Expr}{expr}
\SetKwData{D}{d}
\SetKwData{HasEligibleChild}{hasEligibleChild}
\SetKwData{DummyVarType}{dummyVarType}
\SetKwData{NodeIsRightType}{exprIsRightType}

\SetKwFunction{NextNode}{NextExpression}
\SetKwFunction{FirstChild}{FirstChild}
\SetKwFunction{TypeOf}{TypeOf}
\SetKwFunction{RewriteReturnExpression}{RewriteReturnExpression}

\SetKwInOut{Input}{Input}
\SetKwInOut{Output}{Output}

\SetKw{Continue}{continue}

\SetKwProg{Fn}{Function}{}{end}

\BlankLine
\Input{The return expression of the comprehension, \Expr}
\Input{A list of induction variables}
\Input{The dummy variable type, \DummyVarType}
\Output{The rewritten return expression}
\Output{A list of declared dummy variables}
\BlankLine
\Expr $\leftarrow$ \FirstChild{\Expr}\;
\While{\Expr has a parent}{
    \If{\Expr does not reference any non-induction variables}{
        \Expr $\leftarrow$ \NextNode{\Expr}\;
        \Continue \;
    }
    
    \HasEligibleChild $\leftarrow \exists$ a descendant \D of \Expr, where $\TypeOf{\D} = \DummyVarType$, and \D references non-induction variables\;
    
    \NodeIsRightType $\leftarrow$ $\TypeOf{\Expr} = \DummyVarType$\;

    \If{$!\HasEligibleChild \And \NodeIsRightType$}{
        replace \Expr with a new dummy variable\;
        \Expr $\leftarrow$ \NextNode{\Expr}\;
        \Continue;
    }

    \If{$!\HasEligibleChild \And !\NodeIsRightType$}{
        walk up the expression tree until \Expr is the right type\;
        replace \Expr with a new dummy variable\;
        \Expr $\leftarrow$ \NextNode{\Expr}\;
        \Continue\;
    }

    \tcp{\Expr has an eligible child}
    \If{\Expr references induction variables}{
    \Expr $\leftarrow$ \FirstChild{\Expr}\;
    \Continue;
    } \Else  {
        replace \Expr with a new dummy variable\;
        \Expr $\leftarrow$ \NextNode{\Expr}
    }

    \Return \Expr and a list of newly declared dummy variables\;
}

\BlankLine

\Fn{\NextNode{\Expr}}{
    \Return the right sibling expression of \Expr if one exists, or its parent expression\;
}
\BlankLine

\Fn{\FirstChild{\Expr}}{
\Return the leftmost sub-expression of \Expr\;
}
\end{algorithm}

Now we turn our focus to the algorithm used to perform this substitution.

An important consideration is which sub-expressions to replace with a dummy
variable. Minimizing the number of dummy variables reduces the number of
variables in the generator model, improving performance. For instance, it is
better to rewrite \texttt{m[i] \% 2 = 0 /\textbackslash\  m[i] \% 3 = 0} as
\texttt{Z} than \texttt{Z1 /\textbackslash\ Z2}. At the same time, expressions
containing induction variables should not be placed inside a dummy variable, so
that we do not remove any static guards. Finally, we require the type of dummy
variables to be equal to the type of the identity element of the comprehension.

We approach this as a tree traversal over the return expression, the algorithm
for which is given in \autoref{alg:rewrite}. For each child expression of the
return expression, we check the variables it references. If it references
non-induction variables only, and is of the right type, we turn it into a dummy
variable. If it also references induction variables, we check to see if there
exists a child expression of the right type containing non-induction variables.
If so, we move our cursor to the first child of the current expression, and
repeat this process.

\subsection{Proof of Validity and Correctness}

We consider our algorithm to be valid if we can convert all well-typed return
expressions into static expressions that do not reference any non-induction
variables. Additionally, we consider it correct if the unrolling process does
not remove any return expressions whose values depend on assignments of model 
variables determined at solve-time.

As dummy variables are unconstrained, introducing a dummy variable does not
change the possible values the induction variables can take. Furthermore, when a
dummy variable removes an induction variable from the return expression it
increases the possible combinations of induction variables, delaying the
evaluation of statically known guards to post expansion partial evaluation or
solve-time. As such, introducing a dummy variable only ever weakens the
constraints on the induction variables, and does not affect correctness.

The backtracking case on lines 14 to 19 applies when a sub-expression
still contains non-induction variables and no other cases apply. Here, the
algorithm traverses up the expression tree until it can find an expression of
the right type to convert into a dummy variable. In the worst case, the
algorithm will backtrack up to the root of the expression tree, focusing on the
entire return expression. As the type of a dummy variable is equal to the
expected type of the operands inside the comprehension's operator, and a
well-typed return expression must be this type, we can always turn the entire
return expression into a dummy variable. Therefore, our algorithm is valid for
all well-typed return expressions.

\subsection{Sketch of the Unrolling Process}

In summary, the solver-aided unrolling process consists of four main steps:

\begin{enumerate}
    \item The induction variables and explicit guards from the comprehension are
    added to the generator model.
    \item The return expression is rewritten to a static guard by substituting
    dynamic sub-expressions with dummy variables (\autoref{subsec:thealg}), and
    added to the generator model.
    \item The generator model is solved to generate valid combinations of
    induction variables. 
    \item Each returned combination is substituted into the original return
    expression to produce the fully expanded set of constraints.
\end{enumerate}

\subsection{Unrolling Non-aggregate Comprehensions}

So far, we have focused on the efficient unrolling of aggregate comprehensions.
However, in \EssencePrime, comprehensions can be used inside any operator with a
matrix argument, whether aggregate or not: for example, inside of
an \texttt{allDiff} constraint.

As our return expression rewriting algorithm only works for aggregate
comprehensions, we skip this step for non-aggregate comprehensions, instead
using a generator model constructed from the induction variables and explicit
comprehension guards only. 

\subsection{Implementation}

We implemented this method in a prototype extension to \Conjure \footnote{available online at \url{https://github.com/conjure-cp/conjure-oxide}}, where generator models are expressed in \Essence and solved by \Minion. Integration with other tools such as \SavileRow and \MiniZinc is straightforward. The method introduces negligible overhead when static guards are explicit and provides substantial gains for models where conditions are implicit or not well-placed.


\begin{figure}
  \centering

  \begin{subfigure}{.7\textwidth}
    \centering
    \includegraphics[width=\textwidth]{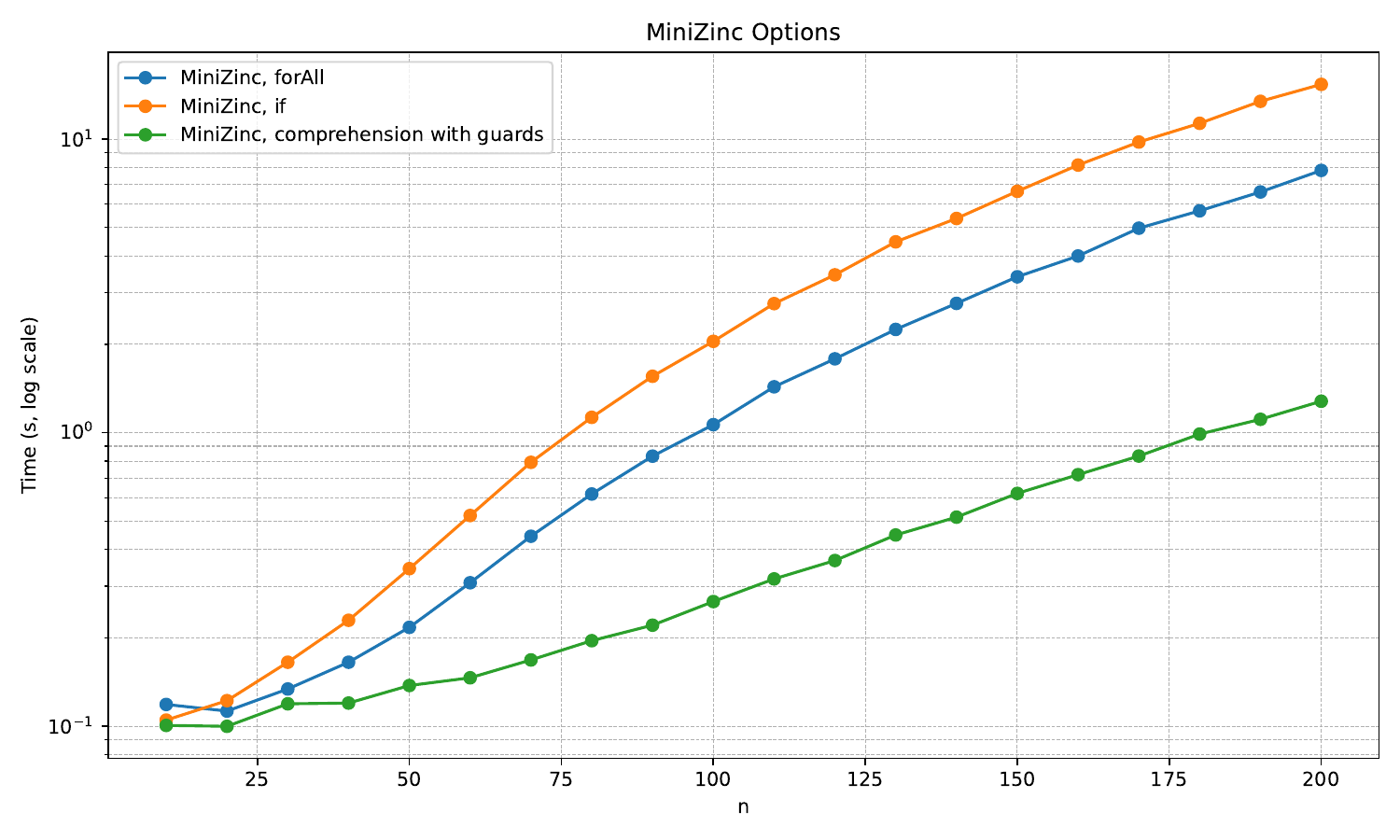}
    \caption{The scaling behaviour of the MiniZinc translations, when targeting Chuffed.}
  \end{subfigure}


  \begin{subfigure}{.7\textwidth}
    \centering
    \includegraphics[width=\textwidth]{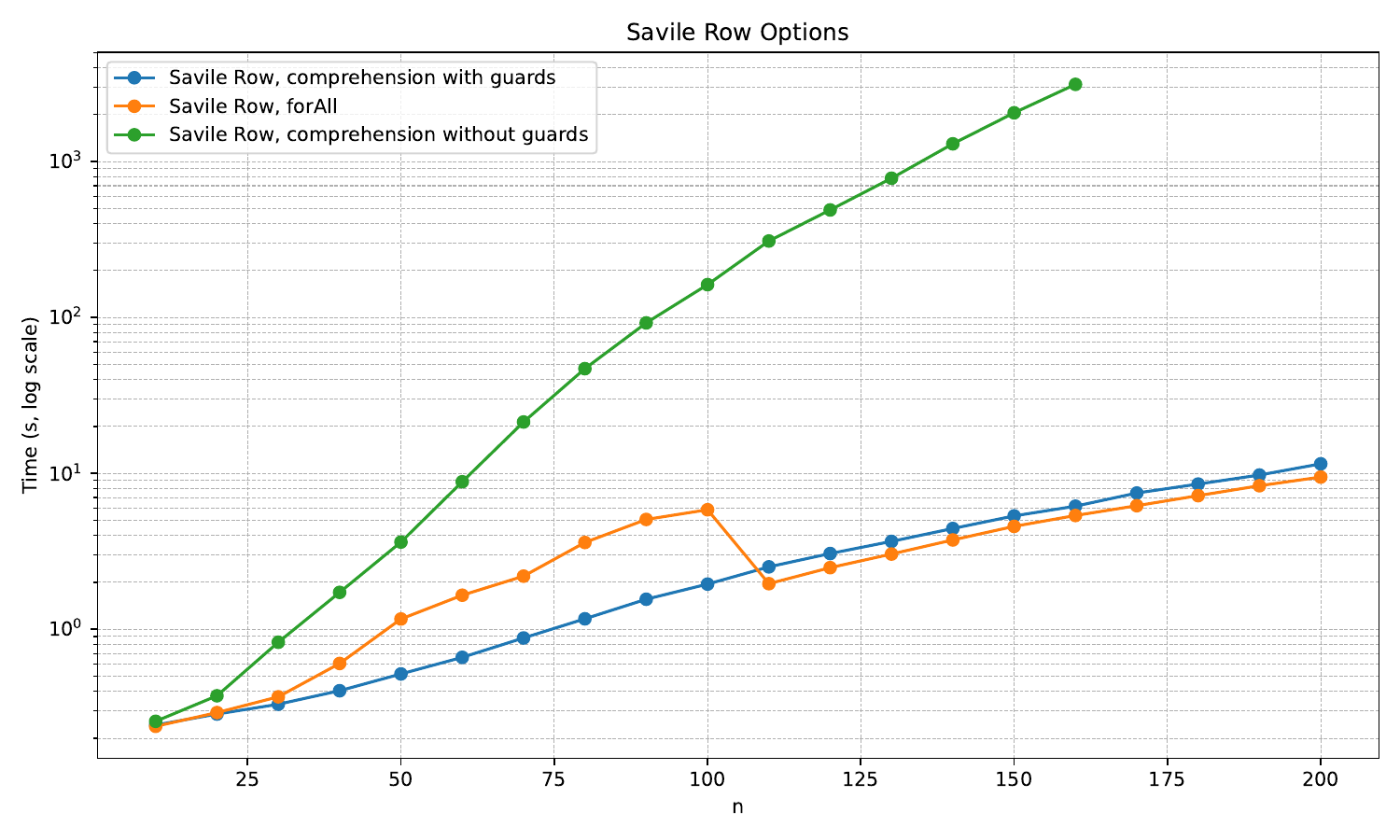}
    \caption{The scaling behaviour of the Savile Row translations, when targeting Minion. 1-hour time limit.}
  \end{subfigure}


  \begin{subfigure}{.7\textwidth}
    \centering
    \includegraphics[width=\textwidth]{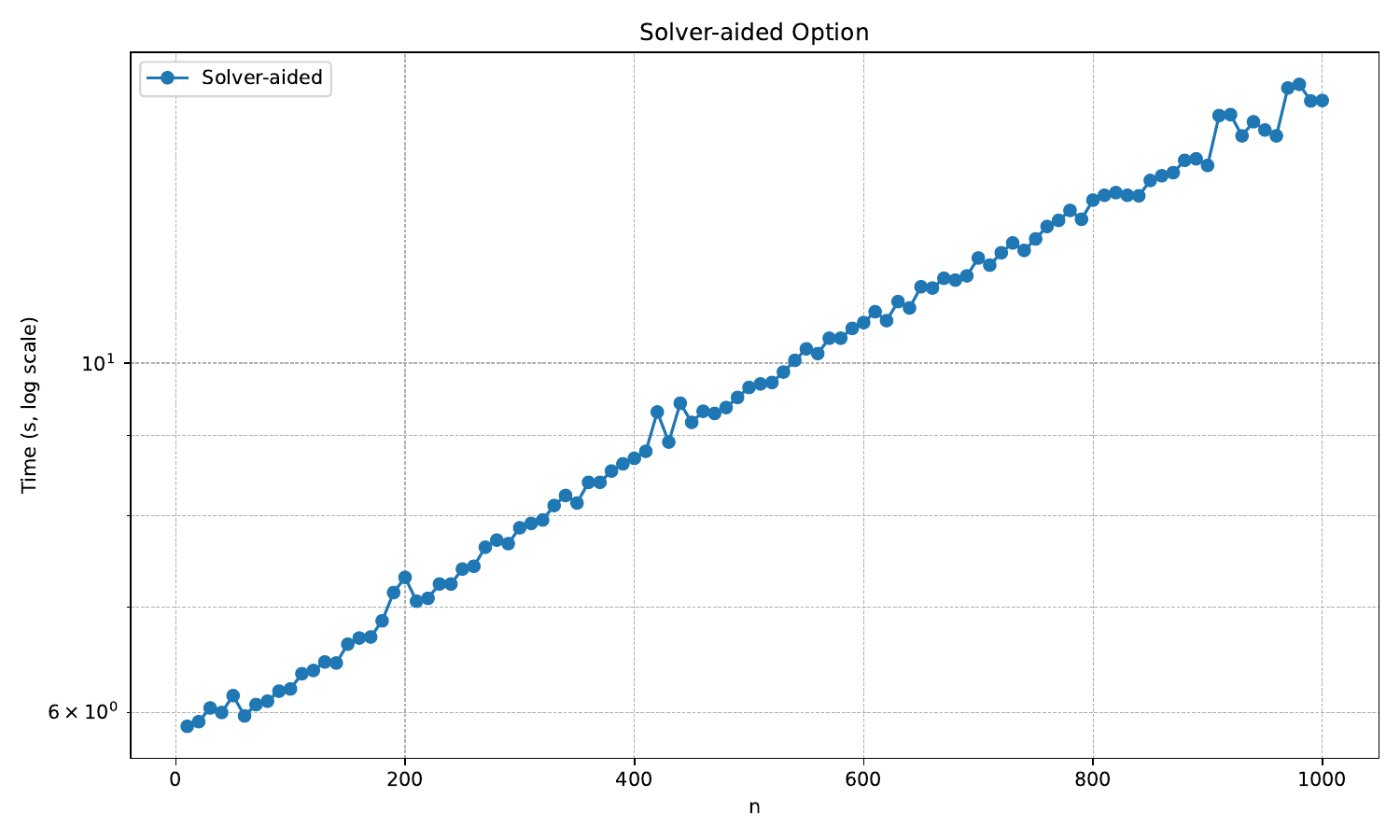}
    \caption{The scaling behaviour of the Solver-aided option. Note that this is a prototype implementation that is not feature-complete. We measure time taken to unroll the input model, without converting it to a target solver.\label{fig:times}}
  \end{subfigure}

  \caption{Comparison of model translation or comprehension expansion times for different modelling approaches. Notice that the solver-aided option ranges from $n=10$ to $n=1000$ whereas the other options range up to $n=200$.}
\end{figure}

\section{Evaluation}
\label{sec:evaluation}

To assess the impact and generality of our solver-aided comprehension expansion technique, we compare its performance against existing approaches in both \MiniZinc{} and \EssencePrime{} (\SavileRow{}) across a range of models and problem sizes. Our focus is on the time taken to expand and flatten comprehensions for constraint solving, as this is the main bottleneck addressed by our approach.

We evaluated the following compilation pipelines:
\begin{itemize}
\item \MiniZinc + Chuffed: Three variants of the Boolean Pythagorean Triples model, differing in how guards are placed (fully explicit comprehension guards, guards embedded in the return expression, and a \texttt{forAll} formulation).
\item \EssencePrime + \SavileRow + \Minion: The same three variants as above.
\item Two variants of our approach: a simple version handling only static guards and a full version that handles both the static guards and introduces dummy variables for dynamic conditions.
\end{itemize}

For each toolchain, we use three representative models of the Boolean Pythagorean Triples problem. These differ only in the syntactic position of guards; whether conditions on the induction variables appear as comprehension guards, are placed inside return expressions, or are handled by quantifiers. This directly affects the performance of current flattening strategies but should not affect our approach, which is designed to be robust to such variation.

We use the model given in \autoref{lst:triples-eprime-inreturnexpr} when running our proposed approaches, which places all guards inside the return expression. This model is particularly problematic for \SavileRow, which must enumerate all variable combinations before partial evaluation, resulting in the slowest compilation times among the methods tested.

\autoref{fig:times} presents the time taken by all eight approaches to expand the models, as $n$ increases. The vertical axis is logarithmic to accommodate the large range of values observed. The figures show that:

\begin{itemize}
\item The time required by \MiniZinc and \SavileRow varies widely depending on how the model is formulated. Poor placement of guards (i.e., inside return expressions rather than as comprehension guards) leads to dramatic increases in time and memory usage.
\item Even the most efficient conventional models scale at least linearly with $n$, but models using generate-and-test can become infeasible at modest problem sizes.
\end{itemize}

An interesting observation is that the \texttt{forAll}-based model for \SavileRow exhibits periodic variation in timing (e.g., $n=120$ is faster than $n=100$), due to the binary splitting algorithm used to expand quantifiers internally.

The models we use in our experiments, scripts we use to run the experiments, raw results and plotting scripts are available in the accompanying repository on GitHub: \url{https://github.com/niklasdewally/2025-comprehension-unrolling-modref}.

The results confirm that current tools for constraint model expansion are sensitive to minor syntactic changes in the model. In contrast, our solver-aided method provides consistent, scalable performance, regardless of how guards and conditions are written.

It is important to note that our prototype is not a full reimplementation of the entire compilation toolchain; the times reported reflect only the comprehension expansion phase. In practice, the benefits observed here would be most pronounced in models where comprehensions dominate the rewriting workload, but even in mixed models the elimination of generate-and-test bottlenecks will improve overall performance.

Our method cannot offer a performance benefit in cases where there are no static conditions or guards to exploit (i.e., when all constraints depend on dynamic variables). In our experiments, for these cases the overhead of constructing and solving a trivial generator model is negligible compared to the baseline. Moreover a dedicated native unrolling algorithm can be employed for these trivial cases to avoid the overheads altogether.


\section{Conclusion}
\label{sec:conclusion}

We introduced a solver-aided approach to comprehension expansion in constraint modelling languages, addressing the inefficiencies associated with conventional generate-and-test methods. By formulating comprehension unrolling as a constraint satisfaction problem, our method systematically extracts and leverages static conditions, thus avoiding the combinatorial explosion typical of current techniques.

Our experimental evaluation demonstrated substantial performance improvements over traditional approaches, especially for models where guards or conditions were implicitly embedded within return expressions. Crucially, the solver-aided approach ensures consistent performance regardless of syntactic variations in the input model, significantly reducing the sensitivity of compilation times to modelling choices.

Future work includes enhancing our prototype to handle a broader set of language constructs, integrating our method fully within popular constraint modelling tools like \SavileRow and \MiniZinc, and exploring solver-aided approaches to other aspects of constraint compilation. Ultimately, our approach moves declarative modelling closer to its ideal: enabling users to focus on clarity and correctness of their models without hidden performance penalties.

\bibliography{refs.bib}

\end{document}